\documentclass[runningheads]{llncs}
\usepackage{graphicx}

\usepackage{tikz}
\usepackage{comment}
\usepackage{amsmath,amssymb} 
\usepackage{color}
\usepackage{multirow}
\usepackage{booktabs}
\usepackage{multirow}
\usepackage{epsfig}
\usepackage{float}
\usepackage{graphics}
\usepackage{float}
\usepackage[accsupp]{axessibility}  

\begin{document}
\pagestyle{headings}
\mainmatter
\def\ECCVSubNumber{00000}  

\title{Mutual Learning for Domain Adaptation: Self-distillation Image Dehazing Network with Sample-cycle} 

\titlerunning{Abbreviated paper title}
%
\author{Tian Ye\inst{1}\orcidID{0000-1111-2222-3333}$^\dag$ \and
Yun Liu$^\dag$\inst{2}\orcidID{1111-2222-3333-4444} \and
Yunchen Zhang$^\dag$\inst{3}\orcidID{2222--3333-4444-5555} \and
Sixiang Chen \inst{1}\orcidID{2222--3333-4444-5555}\and
Erkang Chen\thanks{Corresponding author.$^\dag$Equal contribution.} \inst{1}\orcidID{2222--3333-4444-5555}}
\authorrunning{F. Author et al.}
%
\institute{$^1$School of Ocean Information Engineering,
    Jimei University, Xiamen, China\\
    $^2$ Southwest University School of Artificial Intelligence, Chonqing, China \\
    $^3$China Design Group Co., Ltd.
    Nanjing, China}

\maketitle

\begin{abstract}
Deep learning-based methods have made significant achievements for image dehazing. However, most of existing dehazing networks are concentrated on training models using simulated hazy images, resulting in generalization performance degradation when applied on real-world hazy images because of domain shift. In this paper, we propose a mutual learning dehazing framework for domain adaption. Specifically, we first devise two siamese networks: a teacher network in the synthetic domain and a student network in the real domain, and then optimize them in a mutual learning manner by leveraging EMA and joint loss. 
Moreover, we design a sample-cycle strategy based on density augmentation (HDA) module to introduce pseudo real-world image pairs provided by the student network into training for further improving the generalization performance. Extensive experiments on both synthetic and real-world dataset demonstrate that the propose mutual learning framework outperforms state-of-the-art dehazing techniques in terms of subjective and objective evaluation.

\end{abstract}

\section{Introduction}
Single image dehazing is a significant image processing problem in computer vision community, which aims to recover the clean image from a hazy input. According to the atmospheric scattering model~\cite{narasimhan2003contrast}, the formation of a hazy image is modeled as:
\begin{equation}
I(x)=J(x)t(x)+A(1-t(x))
\end{equation}
where $I(x)$ and $J(x)$ respectively are the observed hazy image and the scene radiance, $A$ denotes the global atmospheric light, and $t(x)$ represents the transmission map indicating the part of light that reaches the camera sensors. Since only an input hazy image $I(x)$ is known, this problem is generally under-constrained.

Early dehazing methods~\cite{he2010single,zhu2015fast,fattal2014,Bu2018Single,Berman2020single,Ju2020IDGCP,Ju2021IDRLP} attempt to remove the haze relying on hand-crafted priors. However, these hand-crated priors from human observation may not always hold in case of the complex real-world hazy scenes. To circumvent hand-crafted priors dependency, numerous deep learning-based networks~\cite{cai2016dehazenet,ren2016single,aod,ren2018gated,kddn,msbdn,ffa-net,wu2021contrastive} have been proposed for image dehazing, which achieves better performance than hand-crafted prior based approaches owing to the powerful feature representation ability of deep convolutional neural networks. As learning-based techniques typically require a large-scale paired data and the real-world image paired data under haze and haze-free conditions is difficult to collect, most of them are focused on training dehazing models with synthetic hazy dataset. Unfortunately, they fail to generalize well to real-world hazy scenes due to the domain gap between the synthetic and real data.

\begin{figure}[t]
\centering
\includegraphics[width=11.5cm]{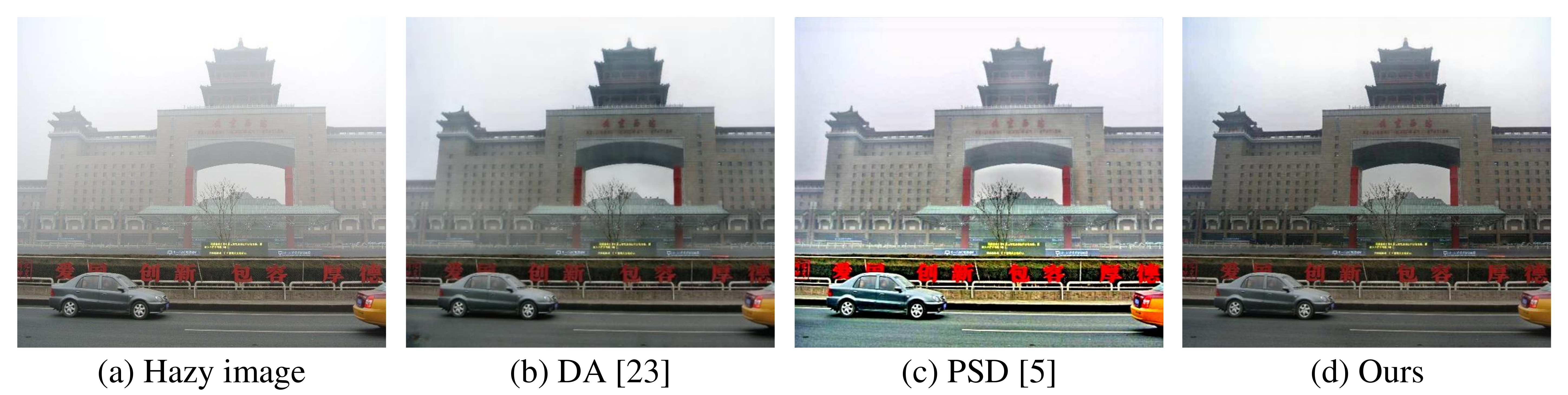}
\caption{Dehazing results on a real-world hazy image. Our method can provide a better dehazing result compared with state-of-the-art methods. }
\label{figs:compairson_real}
\end{figure}

To solve the domain shift problem, some works~\cite{li2019semi,Shao2020Domain,Chen2021PSD} made use of domain translation strategy to reduce the discrepancy between domains. Li~\emph{et al}.~\cite{li2019semi} designed a semi-supervised dehazing framework containing a supervised network and an unsupervised network respectively trained on synthetic and real-world images and these two sub-networks share the weights during the training process. Shao~\emph{et al}.~\cite{Shao2020Domain} developed a domain adaptation paradigm, which consists of a image translation module and two dehazing modules and the image translation module is responsible for bridging the gap between the synthetic and real-world hazy images. Chen~\emph{et al}.~\cite{Chen2021PSD} exploited a pre-trained dehazing model on synthetic images to fine-tune on both synthetic and real hazy images guided by physical priors. These aforementioned approaches leverage weights-sharing or domain translation to build up the relationship between the synthetic and real domains. Nevertheless, as shown in Fig.~\ref{figs:compairson_real}, their dehazing ability on the real domain is insufficient to provide the clean image because they fail to achieve the domain adaptation in essence, resulting in a performance gap between the synthetic and real domains.



In order to achieve the domain adaptation actually, we propose a novel mutual learning paradigm for single image dehazing to address the domain shift problem. The proposed framework includes two siamese networks, namely a teacher network and a student network. First, we train a teacher module in the synthetic domain and transfer useful knowledge from teacher to student. Then, the parameters of the student network in the real domain are updated via the exponential moving average (EMA) and the pseudo haze-free image generated by the student network is evaluated with unsupervised losses in a online manner. Afterwards, the rich external knowledge is constructed by pixel-wise adversarial loss and dark channel prior loss in the real domain and explicitly transferred from student to teacher. Leveraging the external knowledge, the teacher network can teach the student network how to improve the dehazing performance in real domain. In addition, to narrow the gap between synthetic and real domains further, we introduce paired pseudo samples for the teacher network to augment the diversity of haze density. Finally, the obtained student network is capable of yielding pleasant results in the synthetic and real domains simultaneously. As shown in Fig.~\ref{figs:compairson_real}, the obtained student network can yield pleasant dehazing result for a real-world image when compared with other famous domain adaptation methods. The main contributions are summarized as follows:
\begin{itemize}
  \item We propose a mutual learning paradigm for domain adaptation to address the domain gap through knowledge mutual transfer.
  \item We develop a novel data augmentation strategy called haze density augmentation (HDA) in a sample-cycle fashion. The haze-free images acquired by the student network is fed into the HDA module to form the pseudo real-world image pairs with diverse haze density, which is introduced into the teacher network for data augmentation.
  \item We conduct comprehensive experiments on both synthetic datasets and real-world hazy images which verifies the proposed framework performs better than state-of-the-art methods.
\end{itemize}

\section{Related Works}
\subsection{Image Dehazing}
Existing dehazing approaches can be roughly classified into two categories: prior-based methods and learning-based methods. Since single image dehazing is ill-conditioned, early works concentrate on exploiting various human-selected cues from empirical observation, such as dark channel prior (DCP)~\cite{he2010single}, color attenuation prior~\cite{zhu2015fast}, color lines~\cite{fattal2014}, haze-lines~\cite{Berman2020single}, region lines prior~\cite{Ju2021IDRLP} and more, to estimate the atmospheric light and transmission for recovering the haze-free image. Despite of achieving pleasant dehazing results for specific target scenes, these methods cannot satisfy the complex practical scenarios because the priors rely on the relative assumption, resulting in less robustness for real-world scenes. For instance, DCP works not well for the scenes inherently similar to the airlight such as sky regions. Therefore, learning-based approaches make use of the feature extraction capabilities of deep neural networks, such as DehazeNet~\cite{cai2016dehazenet} and MSCNN~\cite{ren2016single}, to estimate the transmission. Recently, several end-to-end methods~\cite{ren2018gated,kddn,msbdn,ffa-net,wu2021contrastive} are presented to directly learn the mapping relationships between hazy inputs and clean images without depending on the atmospheric scattering model. While these approaches have received remarkable performance on dehazing, they are not well equipped to generalize to the real-world hazy images because most of them are trained on synthetic dataset which is not convincing as the synthesized haze cannot faithfully represent the true haze distribution.
\subsection{Domain Adaptation}
In order to alleviate the discrepancy between synthetic and real domain, domain adaptation has been received more and more concern in dehazing task. Some studies~\cite{li2019semi,Shao2020Domain,Chen2021PSD,liu2021synthetic,shyam2021towards} are presented to achieve domain adaptation for increasing the generalization ability on real-world hazy images. For example, Li~\emph{et al}.~\cite{li2019semi} propose a semi-supervised image dehazing (SSID) network consisting of a supervised branch and a unsupervised branch respectively trained on labeled synthetic data and unlabeled real data and utilize weights sharing and unsupervised losses to boost the dehazing performance on real data. Shao~\emph{et al}.~\cite{Shao2020Domain} design the domain adaptation (DA) framework containing two dehazing modules and an image translation module to bridge the domain gap between the synthetic and real-world hazy images. A principle synthetic-to-real dehazing (PSD) framework~\cite{Chen2021PSD} is proposed to attempt to adapt synthetic data based models to the real domain. Liu~\emph{et al}.~\cite{liu2021synthetic} develop a disentangle-consistency mean-teacher network (DMT-Net) collaborating with unlabeled real-world hazy images to address the domain shift problem. An adversarial prior-guided framework~\cite{shyam2021towards} is introduced to achieve domain invariant dehazing by leveraging the frequency prior based dual discriminators. These methods design two or more sub-networks and exploit the weights-sharing or domain translation strategies to achieve domain adaptation, which verifies the effectiveness in boosting the dehazing performance on real data. However, the existing domain splitting paradigms such as SSID, DA and PSD hardly acquire a sub-network that simultaneously works well in synthetic and real domains and these methods typically have less robustness because of relying on the complicated network architecture and physical model. In contrast, we introduce a mutual learning paradigm to transfer knowledge each other between the synthetic domain and real domain online and the final student network guarantees state-of-the-art dehazing performance in both domains.
\section{Proposed Method}
This section presents the details of mutual learning paradigm and self-distillation image dehazing framework. First, we give an overview of our method. Then we describe the details of knowledge transfer and haze density augmentation module. Finally, the loss functions are provided for training the proposed framework.
\begin{figure}
\centering
\includegraphics[width=11.5cm]{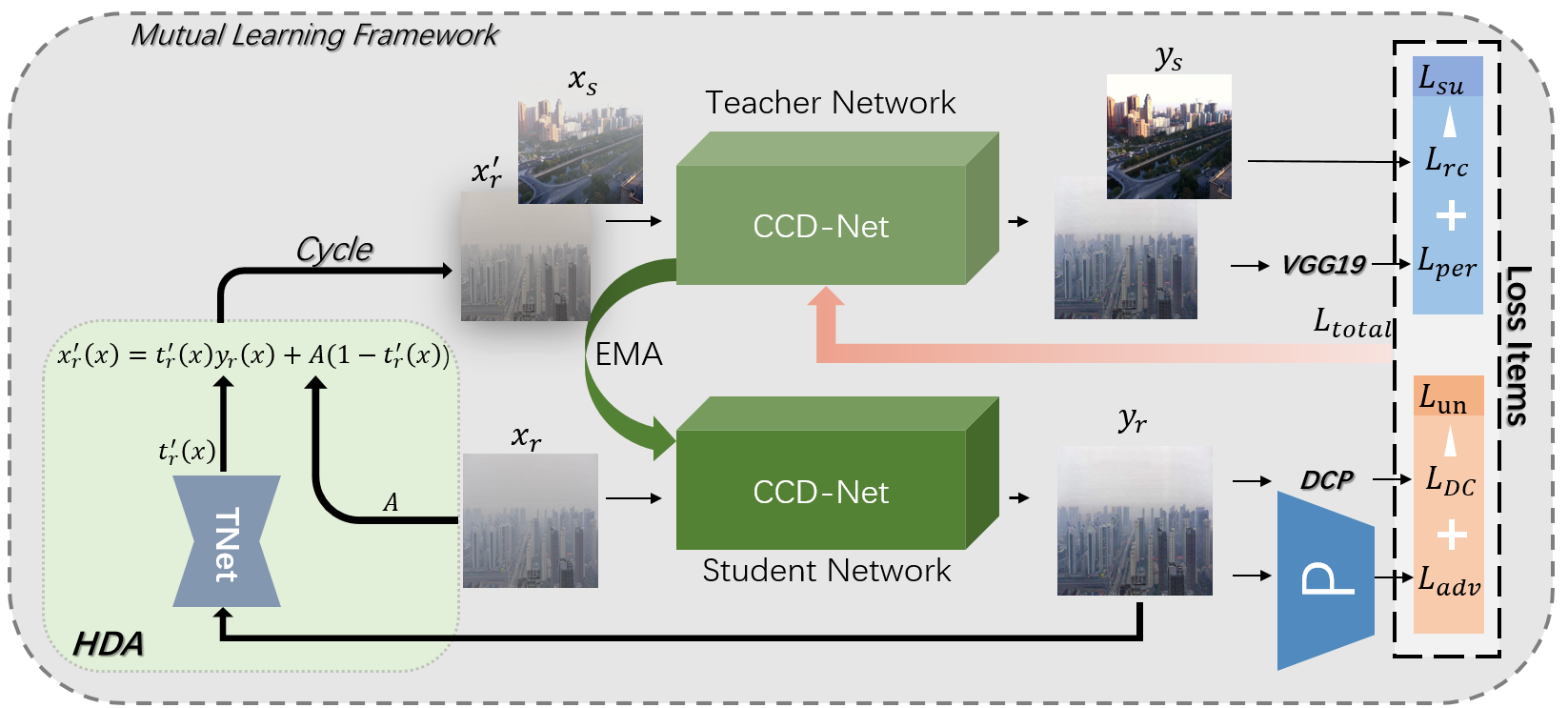}
\caption{Overview of the proposed self-distillation dehazing framework. Our model mainly consists of three parts: a teacher network, a student network and a HDA module. EMA is the exponential moving average. The T-Net of our HDA module estimates a transmission map from the output of student network. The global atmospheric light A is predicted by the DCP. The trapezoid with "P" denotes the discriminator of PatchGAN~\cite{isola2017image}. }
\label{figs:architecture}
\end{figure}

\subsection{Overall framework}
Given a synthetic dataset $X_{syn} = \{{x_i,y_i\}}_{s=1}^{N_s}$ and a real-world hazy dataset $X_{real} = \{{x_i\}}_{r=1}^{N_r}$, where $N_s$ and $N_r$ respectively represent the number of the synthetic and real hazy images. We aim at obtaining a simple and accurate dehazing model which can recover accurately clear images at inferencing phases. We found that most previous dehazing models trained only on the synthetic dataset, and generalized well on the synthetic domain. However, due to the domain gap between the synthetic domain and real domain, these methods 
fail to work well to the image of both domains simultaneously.

To deal with this problem, we present a novel self-distillation image dehazing framework for both domain adaptation, which consists of three main parts: a teacher network $G_{s \rightarrow c}$, a student network $G_{r \rightarrow c}$ and a HDA module $G_{Density}$. With the help of supervised learning, the teacher network learns how to build a mapping from synthetic hazy images to clean images on paired synthetic dataset $X_{syn}$. Under the useful knowledge transferred from $G_{s \rightarrow c}$, the student network learns how to generate satisfactory dehazing results from real-world hazy images. Implicit knowledge transfer considers to transfer the knowledge from teacher to student by weights sharing and joint training, however, which only provides limited internal knowledge.

To fully explore the actual presentation learning ability of the student network, we design a mutual learning mechanism to evaluate the ``real" dehazing performance of the student network online. This end-to-end approach avoids the complicated multi-domain translation training scheme. Moreover, it also enables a ``flywheel effect" that the mutual learning of $G_{s \rightarrow c}$ and $G_{t \rightarrow c}$ can mutually reinforce each other, so that both get better and better as the training goes on.

\subsection{Mutual Learning of T and S}
\begin{figure}[t]
    \centering
    \includegraphics[width=11cm]{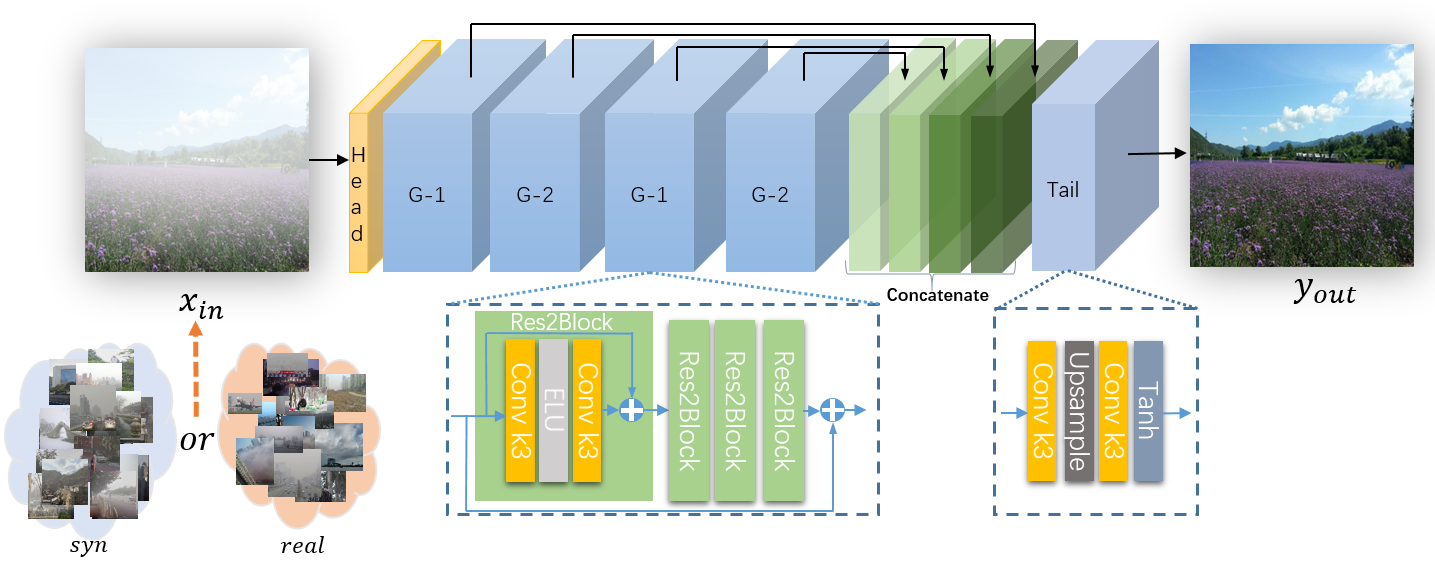}
    \caption{The Compact Cascaded Dehazing Network (CCD-Net) Architecture. }
    \label{fig:NetworkArchitecture}
\end{figure}
Most previous distillation frameworks only transfer implicit knowledge unidirectionally, thus student network only work well in a single domain (e.g, real or synthetic). Why not let the teacher network understand the actual real performance of the student network on the other domain online and know how to teach the student better? The proposed mutual learning mechanism boosts the well performance of student on both domain with the help of mutual learning of T and S, which means the final inferencing network (student) generalize well on both real and synthetic domain, different from the previous domain-splitting adaptation method for image dehazing.

The proposed mutual learning paradigm allows unsupervised loss function to online evaluate the actual performance of the student in the real domain. Specifically, unlabeled real-world hazy images $x_{in}^{r}$ as a batch are randomly sampled from $X_{real}$. The student model is applied on these images by forwarding propagation, thus capturing the adversarial and dark channel prior loss as the
``online evaluated errors":
\begin{equation}
    \mathcal{L}_{un} = \mathcal{L}_{adv}(G_{r \rightarrow c}(x_{in}^r))+\mathcal{L}_{DCP}(G_{r \rightarrow c}(x_{in}^r)).
\end{equation}

Afterwards, $\mathcal{L}_{un}$ will be feedback to the teacher network  $G_{s \rightarrow c}$ for back propagation, which effectively helps the teacher network to teach the student network how to generalize well on real-world hazy images and provides external knowledge to teacher. The mutual learning paradigm explores the closed upper and lower bound constraints of the student network in solution space, which significantly improves the performance of $G_{r\rightarrow c}.$

\subsection{Homogeneous Architecture for Self-distillation Network}
Most previous dehazing approaches~\cite{msbdn,kddn,liu2020trident,liu2021synthetic} attempt to improve the performance by complicated network design with heavy parameters, resulting in difficult deployment. Thus, we aim to devise a compact cascaded dehazing network (CCD-Net) as our distillation network, which only has about 4M parameters, far less than previous prevalent methods, such as MSBDN (31.35M)~\cite{msbdn}, KDDN (5.99M)~\cite{kddn}, TDN (about 48M)~\cite{liu2020trident} and DMT-Net (51.79M)~\cite{liu2021synthetic}.

To this end, we utilize 4 Res2Block groups with 64 channels to extract the features after the down-sample by Head module, which is the convolution operation with the kernel size of $3\times3$ and stride of 2. In order to fully make use of features from different depth of network, we conduct $3\times3$ convolution and concatenation operation of channel dimension to reduce the dimension to 64 channel in the feature aggregation stage. As shown in Fig.~\ref{fig:NetworkArchitecture}, we utilize the pixel shuffle operation to up-sample the features to the original resolution, and we design the Tail module, which fusion the extracted features by $3 \times 3$ convolution. We use the tanh function as the activation of the output to generate final image. We can observe from Fig.~\ref{fig:NetworkArchitecture} and Table~\ref{tabs:comparisons_SOTA} that the proposed self-distillation framework avoids complicated hand-craft on purpose but still works well in the synthetic and real domain.
\begin{figure}
    \setlength{\abovecaptionskip}{0.1cm} 
    \setlength{\belowcaptionskip}{-0.5cm}
    \centering
    \includegraphics[width=11cm]{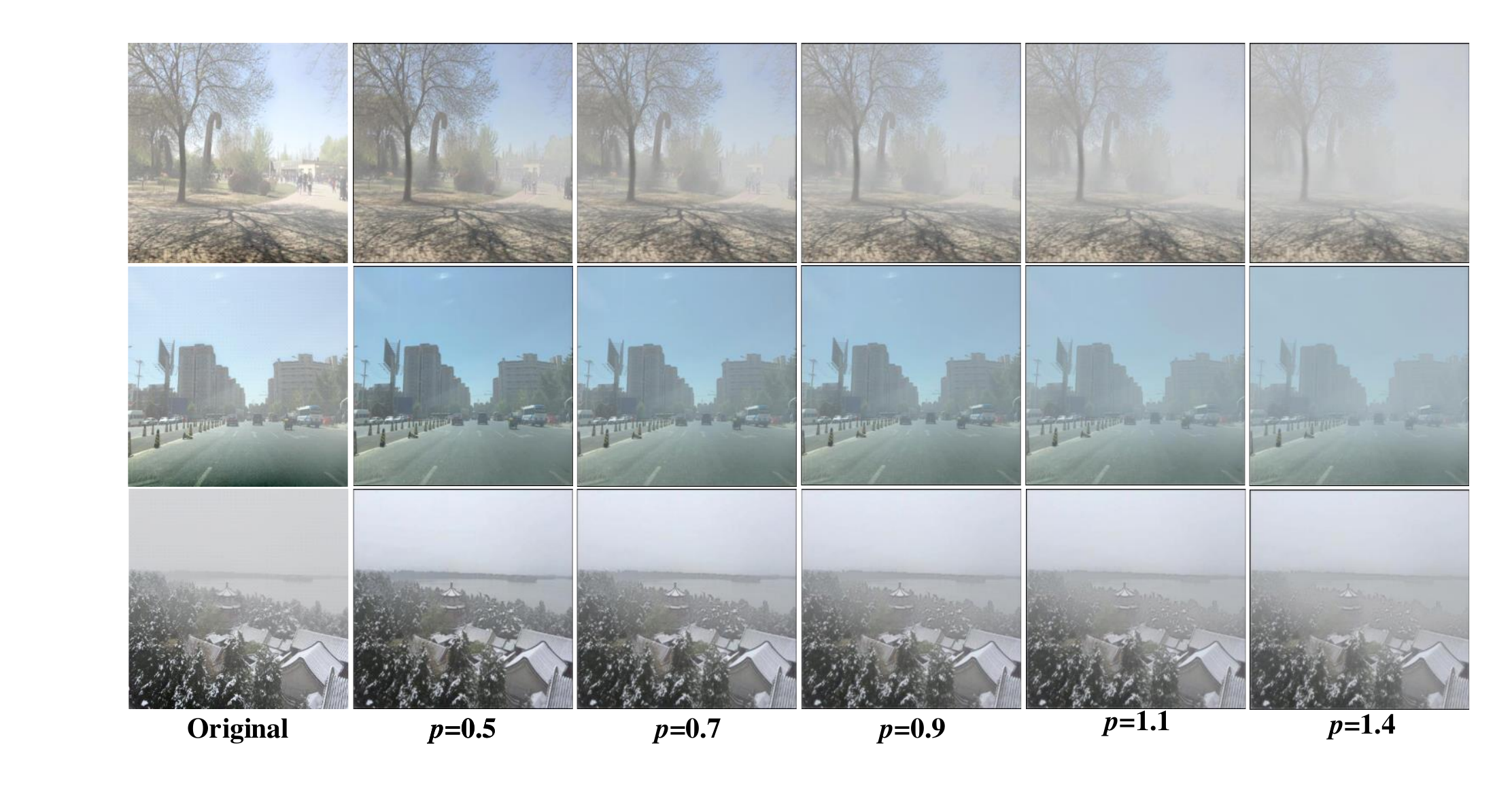}
    \caption{Examples of our HDA module. The images of leftmost column are original hazy images. Zoom in for best view. Please refer to the supplementary materials for more examples of our HDA module on real-world hazy images. }
    \label{figs:HDA_examples}
\end{figure}

\subsection{Sample-cycle with Haze Density Augmentation}
In order to further improve the generalization ability of the student network and enrich the sample complexity of teacher for better domain compatibility, we propose a HDA module to adjust the haze density of real-world images effectively. Different from previous data augmentation for image dehazing methods such as random crop size and random flip which only perform on synthetic datasets or original training samples, we focus on the nonlinear adjustment of transmission in real-world samples for harder hazy samples.

Specifically, we suppose that the real-world dehazed images produced by the student network are clear and satisfactory enough after certain training epochs, thus we can reserve the real-world hazy images and dehazed results to compose the paired samples: $\{ x_r,y_r \}_{r=1}^N$, then we utilize the pre-trained transmission network $T_{r \rightarrow t}$ to estimate the tranmission of $y_r$ accurately:
\begin{equation}
    t_{r}(x) = T_{r \rightarrow t}(y_r)
\end{equation}

To nonlinear augment the hazy density controllably, we set the random factor $p$ to adjust the intensity of transmission and control $t^{'}_{r}(x)$ in $[0,0.99]$:
\begin{equation}
    t^{'}_{r}(x) = t_{r}^p(x_r)
\end{equation}

Finally, we utilize the atmospheric scattering model to rebuild the hazy images with diverse haze density:
\begin{equation}
    x^{'}_r(x) =  t^{'}_{r}(x) y_r(x) + A(1-t^{'}_{r}(x))
\end{equation}
where $A$ is the atmospheric light estimated by the DCP~\cite{he2010single}, and $T_{r \rightarrow t}$ is optimized ceaselessly during the training phrase. Our HDA module effectively rebuilds real-world hazy images and adjusts the haze density controllably, which significantly enrichs the training samples for the teacher network. As observed in Fig.~\ref{figs:HDA_examples}, we give some examples with different haze density generated by our proposed HDA module via adjusting the factor $p$. We found that the most proper range of $p$ is $[0.5, 1.4]$. And we can observe that the HDA module not only enhance the density of original hazy image, but also provides more ``soft" samples for better convergence in training phase. Please refer to the supplementary materials for more detail and ablation study about our HDA module.

After augmentation of haze density, we re-compose $x^{'}_r(x)$ and $y_r(x)$ to form paired training samples. These paired pseudo real-world samples $\{ x^{'}_r,y_r \}$ are conductive for the teacher network to further explore the useful knowledge, which can help the teacher network teach the student network better. In the training stage, the re-composed paired samples are randomly selected to replace the synthetic samples of a batch and the above process likes a special cycle with samples that boosts the generalization ability of the student network.
\subsection{Training Losses}
In the proposed self-distillation dehazing framework, we adopt the following losses to train the network.

~\textbf{Reconstruction Loss}: We utilize the Charbonnier loss \cite{charbonnier1994two} as our reconstruction loss function:
\begin{equation}
    \mathcal{L}_{rc} = \mathcal{L}_{char}(G_{s \rightarrow c}(x_{in}^s),y_{gt}^s) +\lambda_{HDA} \mathcal{L}_{char}(G_{s \rightarrow c}(x_{r}^{'}),y_{r}) .
\end{equation}
where $y_{gt}^s$ stands for ground truth, $\lambda_{HDA}$ is the balance weight for controlling the contribution of the augmented samples, and $\mathcal{L}_{char}$ can be presented by:
\begin{equation}
\mathcal{L}_{\text {char }}=\frac{1}{N} \sum_{i=1}^{N} \sqrt{\left\|X^{i}-Y^{i}\right\|^{2}+\epsilon^{2}}.
\end{equation}
where the constant $\epsilon$ is empirically set as $1e-3$ in our experiments.

~\textbf{Perceptual Loss}: Besides the pixel-wise supervision, we also utilize the perceptual loss based on the VGG-19~\cite{simonyan2014very} pre-trained on ImageNet~\cite{deng2009imagenet}:
\begin{equation}
    \mathcal{L}_{\text {per }}=\sum_{j=1}^{3} \frac{1}{C_{j} H_{j} W_{j}}\left\|\phi_{j}\left(y^{s}_{gt}\right)-\phi_{j}(G_{r \rightarrow c}(x_{in}^s))\right\|_{2}^{2}.
\end{equation}
where $H_j$, $W_j$ and $C_j$ denote the height, width, and channel of the feature map in the j-th layer of the backbone network, $\phi_j$ is the activation of the j-th layer.

~\textbf{Adversarial Loss}: The avdversarial loss $\mathcal{L}_{adv}$ is defined based on the probabilities of the discriminator $D(G_{r\rightarrow c}(x_{r}))$ over the real-world hazy samples as:
\begin{equation}
    \mathcal{L}_{a d v}=\sum_{n=1}^{N}-\log D\left(G_{r\rightarrow c}\left(x^{r}_{in}\right)\right).
\end{equation}

~\textbf{Dark Channel Prior Loss}: The dark channel~\cite{he2010single} is formulated as:
\begin{equation}
DC(x_{in})=\min _{y \in N(x)}\left[\min _{c \in\{r, g, b\}} x_r^{c}(y)\right].
\end{equation}
where $x$ and $y$ are pixel coordinates of image $x_{r}^c$, $x^c$ denotes $c$-th color channel of $x_{r}$, and $N(x)$ represents the local neighbor window centered at $x_{r}$. Previous work~\cite{he2010single} has demonstrated that the most intensity of the dark channel image are close to zero. Thus, we utilize the dark channel loss to enforce that the dark channel of the dehazed images are in consistence with that of haze-free images:
\begin{equation}
\mathcal{L}_{DC}=\left\|DC\left(G_{r \rightarrow c}(x_{in})\right)\right\|_{1}.
\end{equation}

~\textbf{Overall Loss Function}: With all the four members, the overall loss function is defined as follow:
\begin{equation}
    \mathcal{L} = \lambda_{rc} \mathcal{L}_{rc} + \lambda_{adv} \mathcal{L}_{adv}+ \lambda_{DC} \mathcal{L}_{DC} + \lambda_{per} \mathcal{L}_{per}
\end{equation}
where  $\lambda_{rc},\lambda_{adv}, \lambda_{DC}, \lambda_{per}$ are trade-off weights.

\section{Experiments}
In this section, we first describe the implementation details of the proposed framework. Then, the synthetic datasets and real-world images are utilized to evaluate the proposed approach qualitatively and quantitatively. Finally, we conduct the ablation studies to demonstrate the effectiveness of our contributions.

\subsection{Implementation Details}
\textbf{Datasets.} The synthetic and real-world images from RESIDE dataset~\cite{SOTS} are selected for training. The RESIDE~\cite{SOTS} includes five subsets, i.e. ITS (Indoor Training Set), OTS (Outdoor Training Set), SOTS (Synthetic Object Testing Set), URHI (Unannoteated Real Hazy Images) and RTTS (Real Task-driven Testing Set). Concretely, 6000 paired synthetic hazy images, 3000 from ITS and 3000 from OTS, are chose for training. In addition, following~\cite{Shao2020Domain,liu2021synthetic}, we also randomly select 1000 images from URHI in the training stage for improving the generalization ability to the real domain. In order to evaluate the effectiveness of the proposed framework, we compare the proposed framework with other state-of-the-art dehazing methods on two synthetic datasets, i.e. SOTS~\cite{SOTS} and Haze4k~\cite{liu2021synthetic}, and two real-world datasets, i.e. URHI~\cite{SOTS} and I-HAZE~\cite{I-HAZE_2018}. Haze4k~\cite{liu2021synthetic} is a synthesized dataset with 4,000 hazy images, in which the training dataset contains 3,000 paired images (hazy images, transmission maps and clean images) and the testing dataset consists of 1,000 paired images. I-HAZE~\cite{I-HAZE_2018} contains 30 paired real hazy images, in which each hazy image has the associate ground truths of a latent clean image. 
\\
\textbf{Metrics.} We choose two well-known full-reference metrics: Peak Signal-to-Noise Ration (PSNR) and Structural Similarity (SSIM) for comparison quantitatively.
\\
\textbf{Training details.} We augment the training dataset with randomly rotated by 90,180,270 degrees and horizontal flip. And randomly crop all the synthetic and real images to $256\times256$ for the training. We utilize Adam optimizer with initial learning rate of $1\times 10^{-4}$, and adopt the CyclicLR to adjust the learning rate in training, where on the mode of triangular, the value of gamma is 1.0, base momentum is 0.8, max momentum is 0.9, base learning rate is initial learning rate and max learning rate is $1.5\times10^{-4}$. We implement our framework using Pytorch on four GTX 3080 GPU with batch size of 32, which means the usage of 16 synthetic hazy images and 16 real hazy images in each epoch. The decay parameter of EMA is 0.999. The trade-off weights are set as: $\lambda_{rc} = 1$,$\lambda_{adv} = 0.2$,$\lambda_{DC}=10^{-2}$,$\lambda_{per}=0.2$ and $\lambda_{HDA}=0.5$. And we set the patch of as $25\times25$ when computing the dark channel~\cite{he2010single} in $\mathcal{L}_{DC}$.
\\
~\textbf{Inference.} In the testing phase, we only feed the hazy image into the student network and treat the generated predicted image from the student network as the final dehazed result of our method.

\subsection{Comparisons on Synthetic Hazy Images}
\begin{figure}
\centering
\setlength{\abovecaptionskip}{0.1cm} 
\setlength{\belowcaptionskip}{-0.5cm}
\includegraphics[width=12.5cm]{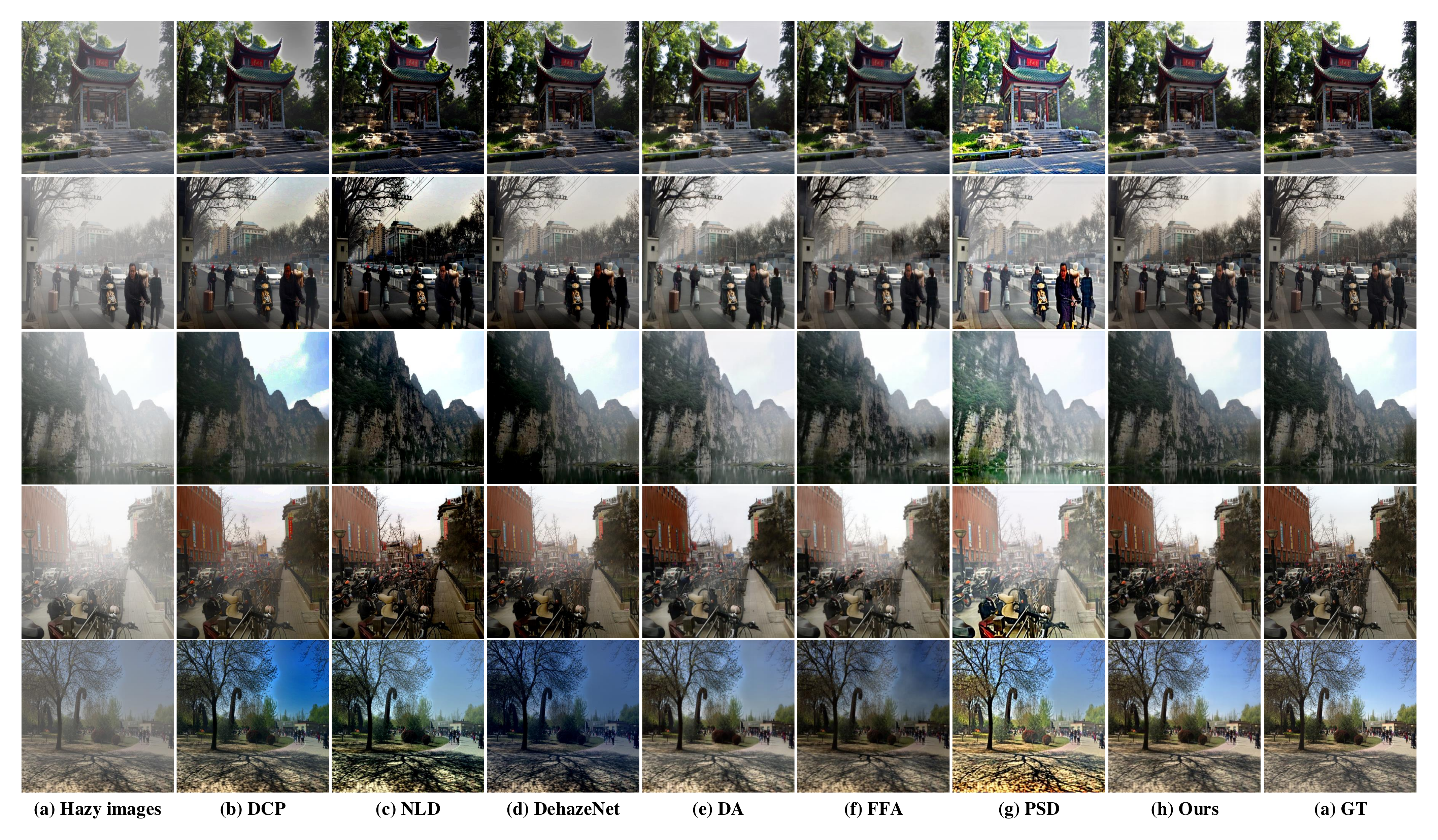}
\caption{Visual comparisons on synthetic hazy images from Haze4k~\cite{liu2021synthetic}.}
\label{figs:comparisons_synthetic}
\end{figure}

The visual comparisons on synthetic hazy images from Haze4k~\cite{liu2021synthetic} are shown in Fig.~\ref{figs:comparisons_synthetic}. From Fig.~\ref{figs:comparisons_synthetic}, traditional prior-based dehazing methods such as DCP~\cite{he2010single}  NLD~\cite{berman2016non} may suffer from the obvious halo artifacts nearby depth discontinuity regions. There still exists some remaining hazes in the results of DehazeNet~\cite{cai2016dehazenet} and FFA~\cite{ffa-net} because of its insufficient dehazing ability. Though the recent domain adaptation methods, i.e. DA~\cite{Shao2020Domain} and PSD~\cite{Chen2021PSD}, can improve the quality of degraded images, but the obtained dehazed results look unrealistic. Compared with the above algorithms, our method can provide better dehazed results in terms of visual effects, which are closer to the ground truths.

In order to avoid the deviation of partial test images, we also provide the quantitative comparisons with several state-of-the-art dehazing methods, including DCP~\cite{he2010single}, NLD~\cite{berman2016non}, DehazeNet~\cite{cai2016dehazenet}, AOD-Net~\cite{aod}, GDN~\cite{griddehazenet}, MSBDN~\cite{msbdn}, DA~\cite{Shao2020Domain},  FFA-Net~\cite{ffa-net}, DID-Net~\cite{liu2021synthetic}, and DMT-Net~\cite{liu2021synthetic}, on two synthetic datasets, namely Haze4k~\cite{liu2021synthetic} and SOTS~\cite{SOTS}, as viewed in Table~\ref{tabs:comparisons_SOTA}. By analyzing Table~\ref{tabs:comparisons_SOTA}, our approach can achieve best scores of PSNR and SSIM in most cases, which demonstrates the superior performance on synthetic hazy images.
\begin{table}[H]
		\centering
			\caption{Quantitative comparisons with the state-of-the-art dehazing methods on three image dehazing datasets (PSNR(dB)/SSIM). Best results are \textbf{bold}. Second-best results are \underline{underlined}.}
			\label{tab:comparisons}
		\resizebox{7.5cm}{!}{
		   \begin{tabular}{l|c|c|c|c|c|c}
		    \toprule[1.2 pt]
			\multirow{2}{*}{Method}             &      \multicolumn{2}{c|}{Haze4k \cite{liu2021synthetic}}       &       \multicolumn{2}{c|}{SOTS \cite{SOTS}}   &       \multicolumn{2}{c}{I-HAZE \cite{I-HAZE_2018}}      \\[1pt]
                      & {PSNR$\uparrow$}  & {SSIM$\uparrow$}  & {PSNR$\uparrow$}  & {SSIM$\uparrow$} & {PSNR$\uparrow$}  & {SSIM$\uparrow$} \\[1pt] \hline
		
			(TPAMI'10)DCP \cite{he2010single}  &       14.01       &       0.76       &         15.09         &0.76  &14.43 &0.75     \\[1pt]
			
			(CVPR'16)NLD\cite{berman2016non}     &15.27 &0.67 &17.27 &0.75 &14.12 &0.65 \\[1pt]
			
			(TIP'16)DehazeNet \cite{cai2016dehazenet}     &       19.12       &       0.84       &       21.14       &       0.85   &10.10 &0.60    \\[1pt]
			(ICCV'17)AOD-Net \cite{aod}                    &       17.15       &       0.83       &  19.06       &       0.85  &13.98 &0.73     \\[1pt]
			(ICCV'19)GDN \cite{griddehazenet}     & 23.29         &         0.93         &  23.29      &       0.95  &16.62 &\underline{0.78}     \\[1pt]
			(CVPR'20)DA\cite{Shao2020Domain}       &24.61              &       0.90       & 27.76       &0.93   &\underline{17.73} &0.73    \\[1pt]
			(AAAI'20)FFA-Net \cite{ffa-net}                  &       26.96       &       0.95       &      26.88       &0.95   &12.70 &0.54    \\[1pt]
			(CVPR'21)PSD-FFA                   &16.56      &0.73      & 13.99   &0.697 &15.04 &0.69      \\[1pt]
			(ACMMM'21)DID-Net \cite{liu2021synthetic}                   &       27.81       &       0.95       &28.30       &0.95 &- &-      \\[1pt] 
			(ACMMM'21)DMT-Net \cite{liu2021synthetic}                   &       \underline{28.53}       &       \underline{0.96}       &\textbf{29.42}       &\textbf{0.97}
			&- &-
			\\[1pt] \hline
			\textbf{Ours}   & \textbf{28.99} & \textbf{0.97} &\underline{28.84}       & \textbf{0.97} &\textbf{18.42} &\textbf{0.81}
		
			\\[1pt] 
			\bottomrule[1.2 pt]
		\end{tabular}
		}
\label{tabs:comparisons_SOTA}
\end{table}
\subsection{Comparisons on Real-world Hazy Images}
\begin{figure}
\setlength{\abovecaptionskip}{0.0cm} 
\setlength{\belowcaptionskip}{-0.5cm}
\centering
\includegraphics[width=12.5cm]{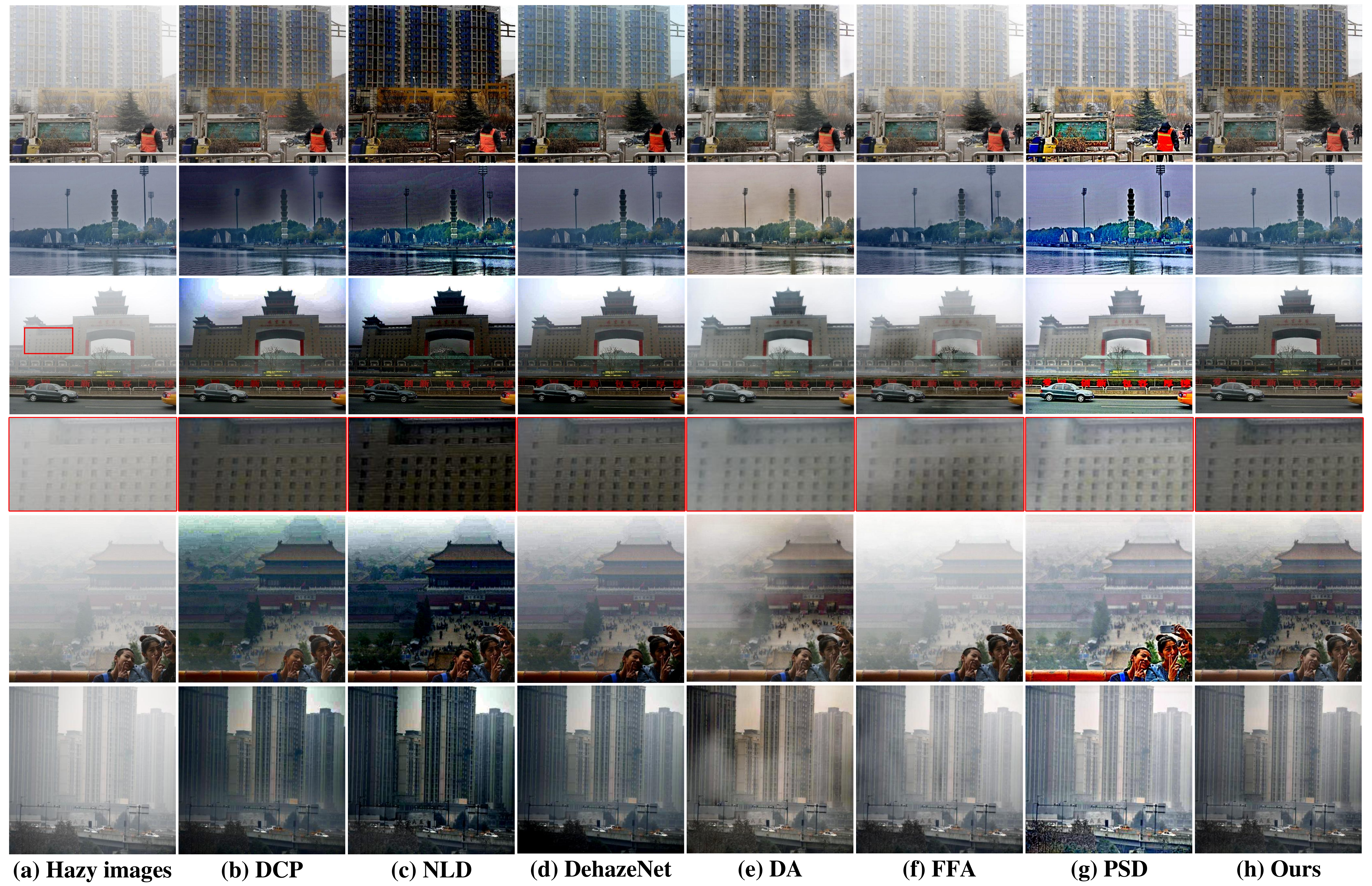}
\caption{Visual comparisons on real-world hazy images from URHI~\cite{SOTS}.}
\label{figs:comparisons_realworld}
\end{figure}

In order to evaluate the generalization ability on the real domain, the comparisons on real-world hazy images selected from URHI dataset~\cite{SOTS} are depicted in Fig.~\ref{figs:comparisons_realworld}. We can observe from Fig.~\ref{figs:comparisons_realworld} that DCP~\cite{he2010single} and NLD~\cite{berman2016non} severely suffer from color distortion in sky regions and the results generated by prior-based methods look unnatural. When handling dense hazy images, the dehazed results of DehazeNet~\cite{cai2016dehazenet} and FFA~\cite{ffa-net} still have remaining hazes as shown in the fifth row of Fig.~\ref{figs:comparisons_realworld}. The domain adaptation methods, such as DA~\cite{Shao2020Domain} and PSD~\cite{Chen2021PSD}, can improve the dehazing performance on real-world hazy images to some extent, but the obvious haze artifacts still remain in the dehazed results, as seen in the fourth row of Fig.~\ref{figs:comparisons_realworld} and the obtained results from PSD~\cite{Chen2021PSD} have the drawback of over-enhancement in the close range regions. In comparison, our approach provides a visual pleasant dehazed results with more details and color rendition.

\begin{figure}
    \vspace{-0.5cm}
    \setlength{\belowcaptionskip}{-0.5cm}
    \centering
    \includegraphics[width=10cm]{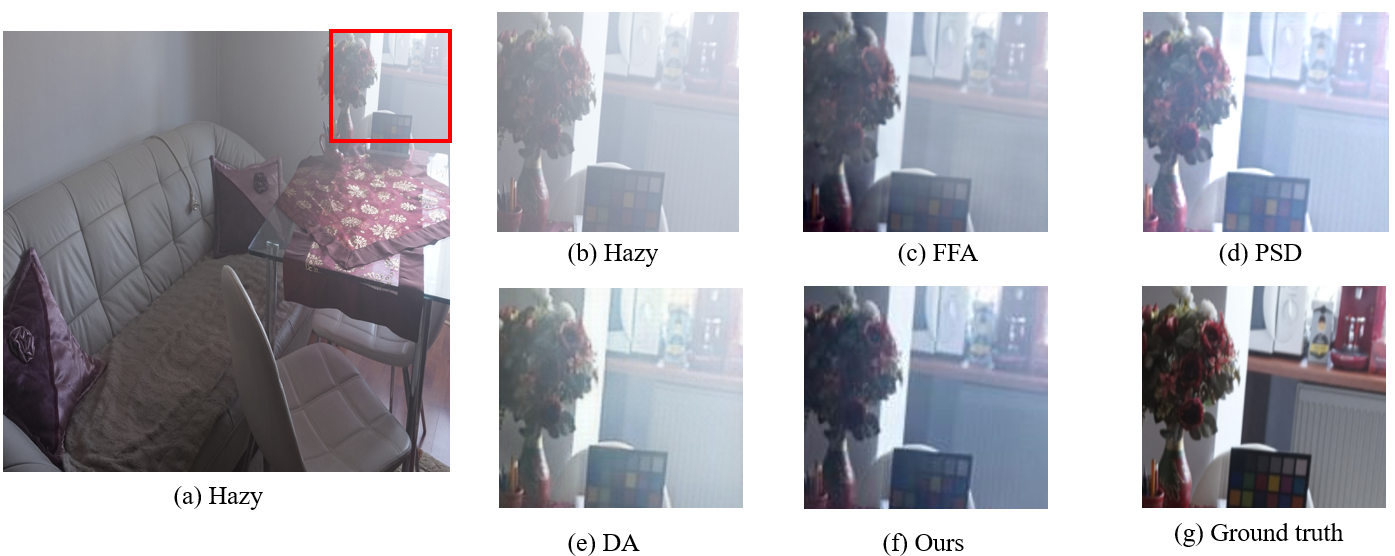}
    \caption{Visual comparison of zoom-in views on a real-world hazy image from I-HAZE dataset~\cite{I-HAZE_2018}. It is worth noting that the training dataset of our method do not contain I-HAZE dataset~\cite{I-HAZE_2018}.}
    \label{figs:comparision_Ihazy_zoom}
\end{figure}

We also give the visual comparison on real-world hazy images from I-HAZE~\cite{I-HAZE_2018}, show in Fig.~\ref{figs:comparision_Ihazy_zoom}.
For a more detailed comparison, Fig.~\ref{figs:comparision_Ihazy_zoom} reveals the zoom-in views of dehazed results provided by FFA~\cite{ffa-net}, PSD~\cite{Chen2021PSD}, DA~\cite{Shao2020Domain} and ours. Overall, our approach can keep more edge details, which are closer to the ground truths. In addition, the real-world hazy dataset namely I-HAZE~\cite{I-HAZE_2018} are tested in Table~\ref{tabs:comparisons_SOTA}. 
We can observe from Table~\ref{tabs:comparisons_SOTA} that our algorithm obtains the highest PSNR and SSIM values for I-HAZE dataset~\cite{I-HAZE_2018}.
\subsection{Ablation Study}
In this section, we follow the same settings of real-world haze training set as described in section 4.1. We choose training set of Haze4k as synthetic training sets. The performance of the network is evaluated on Haze4K test-set and I-Haze test-set. We demonstrate implementation details of the baseline network in the supplementary material. 
\subsubsection{Effectiveness of mutual learning paradigm.} To verify the effectiveness of mutual learning paradigm, we compare our training framework with other self-distillation dehazing frameworks. We conduct experiments on different kinds of training paradigms: 1) single path knowledge transfer from teacher network to student network in the same domain, 2) single path knowledge transfer from teacher network to student network in separated domains, 3) mutual knowledge transfer on the same domain and 4) mutual knowledge transfer on separated domains. The quantitative results are shown in  Table. \ref{tab:ablation_on_training_paradigms}.

\begin{table}[h]
    \vspace{0.2cm}
    \setlength{\abovecaptionskip}{0.1cm} 
    \setlength{\belowcaptionskip}{0.3cm}
	\centering
	\caption{Comparison between different training paradigms of self-distillation dehazing network.}
	\label{tab:ablation_on_training_paradigms}
	{
		\begin{tabular}{ccccc}
		\toprule
		Training Paradigms & Domain of T & Domain of S & $PSNR_{Synthetic}$ & $PSNR_{Real}$ \cr \hline
		T $\rightarrow$ S & Synthetic & Synthetic &27.19 &17.23 \cr
		T $\rightarrow$ S & Synthetic & Real &27.06  &17.82 \cr
		T $\leftrightarrows$ S  & Synthetic & Synthetic &28.05  &17.75 \cr
		T $\leftrightarrows$ S & Synthetic & Real &27.73  &18.03 \cr
		\bottomrule
		\end{tabular}
		}
\end{table}
As is demonstrated in Table. \ref{tab:ablation_on_training_paradigms}, we can see that the single-path knowledge transfer paradigm is inferior to the mutual learning paradigm. 
Specifically, the design of feeding the haze inputs from synthetic domain to the teacher network and feeding the real domain inputs to the student network ensures a better generalization performance in real domain. 
The mutual learning network on the separated domains outperforms the single-path knowledge transfer design with a large margin, while achieving higher scores in both synthetic datasets and real datasets. 
Unfortunately, the domain gap between synthetic and real domains introduces slight performance drop in both single-path knowledge transfer and mutual learning paradigm.

\subsubsection{Effectiveness of Haze Density Augmentations.} Based on the individual results of Table. \ref{tab:ablation_on_training_paradigms}, we verify the effectiveness of the haze density augmentation in narrowing domain gap. 
We follow the design of mutual learning network (last line of Table. \ref{tab:ablation_on_training_paradigms}) and evaluate different settings of haze density augmentations. For a fair comparison, we train the mutual learning network on the same dataset with total epoch of 300.
\begin{table}[h]
	\centering
	\caption{Comparison between different settings of haze density augmentations.}
	\label{tab:ablation_on_haze_density_aug}
	{
		\begin{tabular}{ccccc}
		\toprule
		Training Paradigms & Augmentations & Starting Epochs & $PSNR_{Synthetic}$ & $PSNR_{Real}$ \cr \hline
		T $\leftrightarrows$ S & No & - &27.73  &18.03 \cr
		T $\leftrightarrows$ S & HDA & 0 &28.54  &18.24 \cr
		T $\leftrightarrows$ S & HDA & 20 &28.72  &18.31 \cr
		T $\leftrightarrows$ S & HDA & 50 &28.99  &18.42 \cr
		T $\leftrightarrows$ S & HDA & 80 &28.82  &18.24 \cr
		\bottomrule
		\end{tabular}
		}
\end{table}

Table. \ref{tab:ablation_on_haze_density_aug} shows the performance of different settings of Haze density augmentations. It can be observed that: 1) Haze density augmentation boosts the performance on both synthetic datasets and real-world datasets with a large margin. 2) Different settings of the augmentation strategy influence the performance of overall network. We argue this is because the haze density augmentation adopts an online sampling strategy which directly estimates transmission map from the student network. The inaccurate initialization of the student network results in the unsatisfactory clean output, which interferes with the estimation of transmission map.

\subsection{More Discussions}
~\textbf{Model complexity analysis.} The number of parameters and runtime of our method are 4.02M/0.047s. The model complexity is better than the second-best method (DMT-Net, 51.79M), and the runtime is much worse than the AOD-Net (0.006s). We take the task of accelerating our method and reducing the number of parameters of our framework as one of our future main works.
\\
~\textbf{Performance of our teacher network.} The PSNR/SSIM of our teacher network is 28.46/0.95 on the Haze4k benchmark, worse than our student network with a reduction of 0.53/0.02, which means quantitatively the considerable effectiveness of our mutual learning framework.

\section{Conclusions}
In this paper, we have proposed a novel mutual learning paradigm for domain adaptation. On the basis of this paradigm, our method designs a self-distillation framework consisting of a teacher network and a student network to address the domain shift problem of image dehazing via knowledge mutual transfer within two networks. Moreover, to further improve the generalization ability on the real domain, a novel HDA strategy is presented to augment the haze density of training samples. Experimental results on synthetic datasets and real-world images demonstrate the effectiveness of the proposed framework.

\clearpage
%
%
\bibliographystyle{splncs04}
\bibliography{eccv2022submission}

\begin{thebibliography}{10}
\providecommand{\url}[1]{\texttt{#1}}
\providecommand{\urlprefix}{URL }
\providecommand{\doi}[1]{https://doi.org/#1}

\bibitem{I-HAZE_2018}
Ancuti, C.O., Ancuti, C., Timofte, R., Vleeschouwer, C.D.: I-haze: a dehazing
  benchmark with real hazy and haze-free indoor images. In: arXiv:1804.05091v1
  (2018)

\bibitem{berman2016non}
Berman, D., Avidan, S., et~al.: Non-local image dehazing. In: Proceedings of
  the IEEE conference on computer vision and pattern recognition. pp.
  1674--1682 (2016)

\bibitem{Berman2020single}
Berman, D., Treibitz, T., Avidan, S.: Single image dehazing using haze-lines.
  {IEEE} Trans. Pattern Anal. Mach. Intell.  \textbf{42}(3),  720--734 (2020)

\bibitem{cai2016dehazenet}
Cai, B., Xu, X., Jia, K., Qing, C., Tao, D.: Dehazenet: An end-to-end system
  for single image haze removal. IEEE Transactions on Image Processing
  \textbf{25}(11),  5187--5198 (2016)

\bibitem{charbonnier1994two}
Charbonnier, P., Blanc-Feraud, L., Aubert, G., Barlaud, M.: Two deterministic
  half-quadratic regularization algorithms for computed imaging. In:
  Proceedings of 1st International Conference on Image Processing. vol.~2, pp.
  168--172. IEEE (1994)

\bibitem{Chen2021PSD}
Chen, Z., Wang, Y., Yang, Y., Liu, D.: Psd: Principled synthetic-to-real
  dehazing guided by physical priors. In: Proceedings of the IEEE/CVF
  Conference on Computer Vision and Pattern Recognition (CVPR). pp. 7180--7189
  (June 2021)

\bibitem{deng2009imagenet}
Deng, J., Dong, W., Socher, R., Li, L.J., Li, K., Fei-Fei, L.: Imagenet: A
  large-scale hierarchical image database. In: 2009 IEEE conference on computer
  vision and pattern recognition. pp. 248--255. Ieee (2009)

\bibitem{msbdn}
Dong, H., Pan, J., Xiang, L., Hu, Z., Zhang, X., Wang, F., Yang, M.H.:
  Multi-scale boosted dehazing network with dense feature fusion. In:
  Proceedings of the IEEE/CVF Conference on Computer Vision and Pattern
  Recognition. pp. 2157--2167 (2020)

\bibitem{fattal2014}
Fattal, R.: Dehazing using color-lines. ACM Trans. Graph.  \textbf{34}(1),
  13:1--13:14 (2014)

\bibitem{he2010single}
He, K., Sun, J., Tang, X.: Single image haze removal using dark channel prior.
  IEEE transactions on pattern analysis and machine intelligence
  \textbf{33}(12),  2341--2353 (2010)

\bibitem{kddn}
Hong, M., Xie, Y., Li, C., Qu, Y.: Distilling image dehazing with heterogeneous
  task imitation. In: Proceedings of the IEEE/CVF Conference on Computer Vision
  and Pattern Recognition. pp. 3462--3471 (2020)

\bibitem{isola2017image}
Isola, P., Zhu, J.Y., Zhou, T., Efros, A.A.: Image-to-image translation with
  conditional adversarial networks. In: Proceedings of the IEEE conference on
  computer vision and pattern recognition. pp. 1125--1134 (2017)

\bibitem{Ju2021IDRLP}
Ju, M., Ding, C., Guo, C.A., Ren, W., Tao, D.: {IDRLP:} image dehazing using
  region line prior. {IEEE} Trans. Image Process.  \textbf{30},  9043--9057
  (2021)

\bibitem{Ju2020IDGCP}
Ju, M., Ding, C., Guo, Y.J., Zhang, D.: {IDGCP:} image dehazing based on gamma
  correction prior. {IEEE} Trans. Image Process.  \textbf{29},  3104--3118
  (2020)

\bibitem{aod}
Li, B., Peng, X., Wang, Z., Xu, J., Feng, D.: Aod-net: All-in-one dehazing
  network. In: Proceedings of the IEEE international conference on computer
  vision. pp. 4770--4778 (2017)

\bibitem{SOTS}
Li, B., Ren, W., Fu, D., Tao, D., Feng, D., Zeng, W., Wang, Z.: Benchmarking
  single-image dehazing and beyond. IEEE Transactions on Image Processing
  \textbf{28}(1),  492--505 (2018)

\bibitem{li2019semi}
Li, L., Dong, Y., Ren, W., Pan, J., Gao, C., Song, Y., Ming-Hsuan, Y.:
  Semi-supervised image dehazing. IEEE Transactions on Image Processing
  \textbf{19},  2766--2779 (2019)

\bibitem{liu2020trident}
Liu, J., Wu, H., Xie, Y., Qu, Y., Ma, L.: Trident dehazing network. In:
  Proceedings of the IEEE/CVF Conference on Computer Vision and Pattern
  Recognition Workshops. pp. 430--431 (2020)

\bibitem{griddehazenet}
Liu, X., Ma, Y., Shi, Z., Chen, J.: Griddehazenet: Attention-based multi-scale
  network for image dehazing. In: Proceedings of the IEEE/CVF International
  Conference on Computer Vision. pp. 7314--7323 (2019)

\bibitem{liu2021synthetic}
Liu, Y., Zhu, L., Pei, S., Fu, H., Qin, J., Zhang, Q., Wan, L., Feng, W.: From
  synthetic to real: Image dehazing collaborating with unlabeled real data. In:
  {ACM} Multimedia. pp. 50--58. {ACM} (2021)

\bibitem{narasimhan2003contrast}
Narasimhan, S.G., Nayar, S.K.: Contrast restoration of weather degraded images.
  IEEE transactions on pattern analysis and machine intelligence
  \textbf{25}(6),  713--724 (2003)

\bibitem{ffa-net}
Qin, X., Wang, Z., Bai, Y., Xie, X., Jia, H.: Ffa-net: Feature fusion attention
  network for single image dehazing. In: Proceedings of the AAAI Conference on
  Artificial Intelligence. vol.~34, pp. 11908--11915 (2020)

\bibitem{ren2016single}
Ren, W., Liu, S., Zhang, H., Pan, J., Cao, X., Yang, M.H.: Single image
  dehazing via multi-scale convolutional neural networks. In: Proc. Eur. Conf.
  Comput. Vis. pp. 154--169 (2016)

\bibitem{ren2018gated}
Ren, W., Ma, L., Zhang, J., Pan, J., Cao, X., Liu, W., Yang, M.H.: Gated fusion
  network for single image dehazing. In: Proceedings of the IEEE Conference on
  Computer Vision and Pattern Recognition. pp. 3253--3261 (2018)

\bibitem{Shao2020Domain}
Shao, Y., Li, L., Ren, W., Gao, C., Sang, N.: Domain adaptation for image
  dehazing. In: IEEE/CVF Conference on Computer Vision and Pattern Recognition
  (CVPR) (June 2020)

\bibitem{shyam2021towards}
Shyam, P., Yoon, K., Kim, K.: Towards domain invariant single image dehazing.
  In: {AAAI}. pp. 9657--9665. {AAAI} Press (2021)

\bibitem{simonyan2014very}
Simonyan, K., Zisserman, A.: Very deep convolutional networks for large-scale
  image recognition. arXiv preprint arXiv:1409.1556  (2014)

\bibitem{Bu2018Single}
Trung, B.M., Kim, W.: Single image dehazing using color ellipsoid prior. {IEEE}
  Trans. Image Process.  \textbf{27}(2),  999--1009 (2018)

\bibitem{wu2021contrastive}
Wu, H., Qu, Y., Lin, S., Zhou, J., Qiao, R., Zhang, Z., Xie, Y., Ma, L.:
  Contrastive learning for compact single image dehazing. In: Proceedings of
  the IEEE/CVF Conference on Computer Vision and Pattern Recognition. pp.
  10551--10560 (2021)

\bibitem{zhu2015fast}
Zhu, Q., Mai, J., Shao, L.: A fast single image haze removal algorithm using
  color attenuation prior. IEEE Trans. Image Process.  \textbf{24}(11),
  3522--3533 (Nov 2015)

\end{thebibliography}
\end{document}